\documentclass{article} 
\usepackage{nips15submit_e,times}
\usepackage{url}

\title{Probabilistic Line Searches\\ for Stochastic Optimization}

\author{ 
Maren Mahsereci and Philipp Hennig\\
Max Planck Institute for Intelligent Systems\\
Spemannstra{\ss}e 38, 72076 T\"ubingen, Germany \\
\texttt{[mmahsereci|phennig]@tue.mpg.de}\\
}
 
\usepackage{microtype,marvosym} 
\usepackage{amsmath,amssymb,graphicx}

\usepackage[numbers]{natbib}

\newcommand{\g}{\,|\,} 
\newcommand{\de}{\partial}
 
\newcommand{\eps}{\epsilon}

\newcommand{\Exp}{\mathbb{E}}

\newcommand{\erf}{\operatorname{erf}}

\renewcommand{\Re}{\mathbb{R}}

\newcommand{\N}{\mathcal{N}}

\newcommand{\Trans}{^{\intercal}}

\newcommand{\q}{\quad}
\newcommand{\qq}{\qquad}

\newcommand{\qqqq}{\qquad\qquad}

\renewcommand{\vec}{\boldsymbol}

\newcommand{\GP}{\mathcal{GP}}
\newcommand{\Id}{\vec{I}}

\newcommand{\II}{\mathbb{I}}

\newcommand{\y}{\vec{y}}

\newcommand{\V}{\mathbb{V}}

\usepackage{colonequals}

\usepackage{nicefrac}

\DeclareSymbolFont{stmry}{U}{stmry}{m}{n}
\DeclareMathSymbol\leftarrowtriangle\mathrel{stmry}{"5E}
\DeclareMathSymbol\rightarrowtriangle\mathrel{stmry}{"5F}
\renewcommand{\gets}{\operatorname*{\leftarrowtriangle}}
\renewcommand{\to}{\operatorname*{\rightarrowtriangle}}

\usepackage{tensor}
\newcommand{\dk}{\tensor[^{\de}]{k}{}}
\newcommand{\ddk}{\tensor[^{\de^2}]{k}{}}
\newcommand{\ddkd}{\tensor[^{\de^2}]{k}{^{\de}}}
\newcommand{\dddk}{\tensor[^{\de^3}]{k}{}}
\newcommand{\dddkd}{\tensor[^{\de^3}]{k}{^{\de}}}
\newcommand{\kd}{\tensor{k}{^{\de}}}
\newcommand{\dkd}{\tensor[^{\de}]{k}{^{\de}}}

\usepackage{algorithm,algorithmic}

\newcommand{\pW}{p^\text{Wolfe}}

\newcommand{\sgd}{{\sc sgd}} 

\newcommand{\gp}{{\sc gp}}
\usepackage{pifont}

\newlength\figheight%
\newlength\figwidth%

\nipsfinalcopy 

\begin{document}

\maketitle

\begin{abstract}
In deterministic optimization, line searches are a standard tool ensuring stability and efficiency. Where only stochastic gradients are available, no direct equivalent has so far been formulated, because uncertain gradients do not allow for a strict sequence of decisions collapsing the search space. We construct a probabilistic line search by combining the structure of existing deterministic methods with notions from Bayesian optimization. Our method retains a Gaussian process surrogate of the univariate optimization objective, and uses a probabilistic belief over the Wolfe conditions to monitor the descent. The algorithm has very low computational cost, and no user-controlled parameters. Experiments show that it effectively removes the need to define a learning rate for stochastic gradient descent. 
\end{abstract}

\section{Introduction}
\label{sec:introduction}

Stochastic gradient descent (\sgd)~\cite{robbins1951stochastic} is currently the standard in machine learning for the optimization of highly multivariate functions if their gradient is corrupted by noise. This includes the online or batch training of neural networks, logistic regression~\cite{Zhang:2004,bottou2010large} and variational models \citep[e.g.][]{hoffman2013stochastic,hensman2012fast,StreamingBayes}. In all these cases, noisy gradients arise because an exchangeable loss-function $\mathcal{L}(x)$ of the optimization parameters $x\in\Re^D$, across a large dataset $\{d_i\}_{i=1\,\dots,M}$, is evaluated only on a subset $\{d_j\}_{j=1,\dots,m}$:
\begin{equation}
  \label{eq:1}
  \mathcal{L}(x) := \frac{1}{M}\sum_{i=1} ^M \ell(x,d_i) 
  \approx \frac{1}{m}\sum_{j=1} ^{m} \ell(x,d_j) =: \hat{\mathcal{L}}(x)\qqqq m\ll M.
\end{equation}%
If the indices $j$ are i.i.d.~draws from $[1,M]$, by the Central Limit Theorem, the error $\hat{\mathcal{L}}(x)-\mathcal{L}(x)$ is unbiased and approximately normal distributed. Despite its popularity and its low cost per step, \sgd~has well-known deficiencies that can make it inefficient, or at least tedious to use in practice. Two main issues are that, first, the gradient itself, even without noise, is not the optimal search \emph{direction}; and second, \sgd~requires a \emph{step size} (learning rate) that has drastic effect on the algorithm's efficiency, is often difficult to choose well, and virtually never optimal for each individual descent step. The former issue, adapting the search direction, has been addressed by many authors \citep[see][for an overview]{george2006adaptive}. Existing approaches range from lightweight `diagonal preconditioning' approaches like {\sc adagrad} \cite{duchi2011adaptive} and `stochastic meta-descent'\cite{schraudolph1999local}, to empirical estimates for the natural gradient \cite{amari2000adaptive} or the Newton direction \cite{roux2010fast}, to problem-specific algorithms \cite{ranganath13}, and more elaborate estimates of the Newton direction \cite{StochasticNewton}. Most of these algorithms also include an auxiliary adaptive effect on the learning rate. And Schaul et al.~\cite{schaul2013no} recently provided an estimation method to explicitly adapt the learning rate from one gradient descent step to another. None of these algorithms change the size of the \emph{current} descent step. Accumulating statistics across steps in this fashion requires some conservatism: If the step size is initially too large, or grows too fast, \sgd~can become unstable and `explode', because individual steps are not checked for robustness at the time they are taken.

\begin{figure}[ht]
  \begin{minipage}[c]{0.3\textwidth}
  \centering
  \setlength{\figwidth}{\textwidth}
  \setlength{\figheight}{.18\textheight}
  {\scriptsize \includegraphics{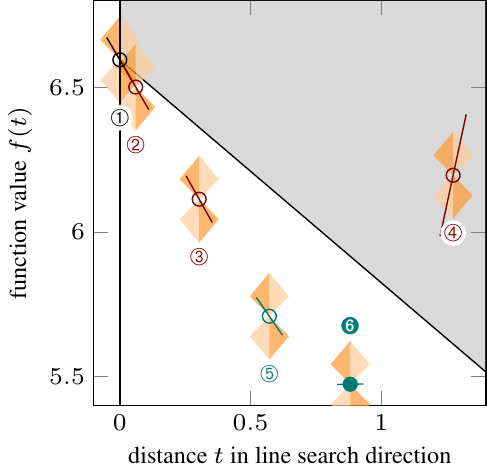}}
  \end{minipage}\hfill
  \begin{minipage}[c]{.6\textwidth}
  \caption{Sketch: The task of a {\bfseries classic line search} is to tune the step taken by a optimization algorithm along a univariate search direction. The search starts at the endpoint \ding{192} of the previous line search, at $t=0$. A sequence of exponentially growing extrapolation steps \ding{193},\ding{194},\ding{195} finds a point of positive gradient at \ding{195}. It is followed by interpolation steps \ding{196},\ding{207} until an acceptable point \ding{207} is found. Points of insufficient decrease, above the line $f(0) + c_1tf'(0)$ (gray area) are excluded by the Armijo condition W-I, while points of steep gradient (orange areas) are excluded by the curvature condition W-II (weak Wolfe conditions in solid orange, strong extension in lighter tone). Point \ding{207} is the first to fulfil both conditions, and is thus accepted.}
\label{fig:minimize-sketch}
  \end{minipage}
\end{figure}
The principally same problem exists in deterministic (noise-free) optimization problems. There, providing stability is one of several tasks of the \emph{line search} subroutine. It is a standard constituent of algorithms like the classic nonlinear conjugate gradient \cite{fletcher1964function} and BFGS \cite{broyden1969new,fletcher1970new,goldfarb1970family,shanno1970conditioning} methods \citep[\textsection 3]{nocedal1999numerical}.\footnote{In these algorithms, another task of the line search is to guarantee certain properties of surrounding estimation rule. In BFGS, e.g., it ensures positive definiteness of the estimate. This aspect will not feature here.} In the noise-free case, line searches are considered a solved problem \citep[\textsection 3]{nocedal1999numerical}. But the methods used in deterministic optimization are not stable to noise. They are easily fooled by even small disturbances, either becoming overly conservative or failing altogether. The reason for this brittleness is that existing line searches take a sequence of hard decisions to shrink or shift the search space. This yields efficiency, but breaks hard in the presence of noise. Section~\ref{sec:method} constructs a probabilistic line search for noisy objectives, stabilizing optimization methods like the works cited above. As line searches only change the length, not the direction of a step, they could be used in combination with the algorithms adapting \sgd's direction, cited above. The algorithm presented below is thus a complement, not a competitor, to these methods.

\section{Connections}
\label{sec:connections}

\subsection{Deterministic Line Searches}
\label{sec:determ-lines}

There is a host of existing line search variants \citep[\textsection 3]{nocedal1999numerical}. In essence, though, these methods explore a univariate domain `to the right' of a starting point, until an `acceptable' point is reached (Figure~\ref{fig:minimize-sketch}). More precisely, consider the problem of minimizing $\mathcal{L}(x):\Re^D\to \Re$, with access to $\nabla \mathcal{L}(x):\Re^D\to\Re^D$. At iteration $i$, some `outer loop' chooses, at location $x_i$, a search direction $s_i\in\Re^D$ (e.g.~by the BFGS rule, or simply $s_i=-\nabla \mathcal{L} (x_i)$ for gradient descent). It will \emph{not} be assumed that $s_i$ has unit norm. The line search operates along the univariate domain $x(t) = x_i + t s_i$ for $t\in\Re_+$. Along this direction it collects scalar function values and projected gradients that will be denoted $f(t) = \mathcal{L} (x(t))$ and $f'(t) = s_i\Trans \nabla \mathcal{L} (x(t))\in\Re$.  Most line searches involve an initial extrapolation phase to find a point $t_r$ with $f'(t_r)>0$. This is followed by a search in $[0,t_r]$, by interval nesting or by interpolation of the collected function and gradient values, e.g.~with cubic splines.\footnote{This is the strategy in {\scriptsize\tt minimize.m} by C.~Rasmussen, which provided a model for our implementation. At the time of writing, it can be found at {\scriptsize\url{http://learning.eng.cam.ac.uk/carl/code/minimize/minimize.m}}}

\subsubsection{The Wolfe Conditions for Termination}
\label{sec:wolfe-cond-conv}

\begin{figure}
\begin{minipage}[c]{.45\textwidth}
  \centering
  \setlength{\figwidth}{.8\textwidth}
  \setlength{\figheight}{.24\textheight}
  {\scriptsize \includegraphics{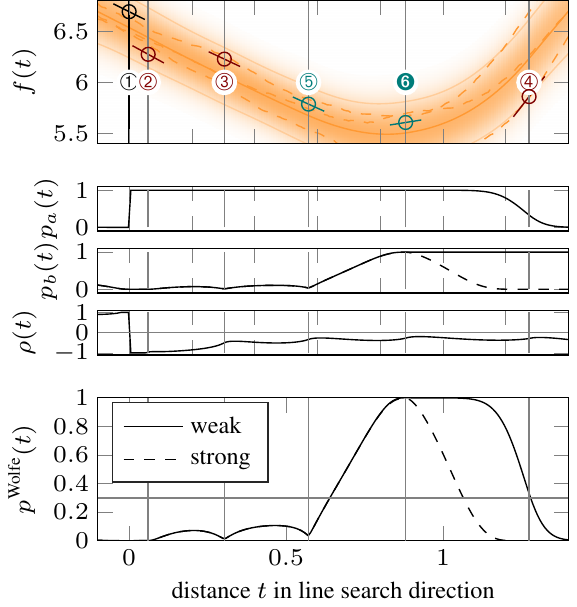}}
\end{minipage}
\begin{minipage}[c]{.55\textwidth}
  \caption{Sketch of a {\bf probabilistic line search}. As in Fig.~\ref{fig:minimize-sketch}, the algorithm performs extrapolation (\ding{193},\ding{194},\ding{195}) and interpolation (\ding{196},\ding{207}), but receives unreliable, noisy function and gradient values. These are used to construct a \gp~posterior (top.~solid posterior mean, thin lines at 2 standard deviations, local pdf marginal as shading, three dashed sample paths). This implies a bivariate Gaussian belief (\textsection\ref{sec:determ-conv}) over the validity of the weak Wolfe conditions (middle three plots. $p_a(t)$ is the marginal for W-I, $p_b(t)$ for W-II, $\rho(t)$ their correlation). Points are considered acceptable if their joint probability $\pW(t)$ (bottom) is above a threshold (gray). An approximation (\textsection\ref{sec:appr-strong-wolfe}) to the strong Wolfe conditions is shown dashed.}
\label{fig:prob_ls_sketch}
\end{minipage}
\end{figure}

As the line search is only an auxiliary step within a larger iteration, it need not find an exact root of $f'$; it suffices to find a point `sufficiently' close to a minimum. The \emph{Wolfe~\cite{wolfe1969convergence} conditions} are a widely accepted formalization of this notion; they consider $t$ acceptable if it fulfills
\begin{equation}
  \label{eq:4}
  f(t) \leq f(0) + c_1 t f'(0)\q\text{(W-I)}\qq\text{and}\qq  f'(t) \geq c_2 f'(0) \q\text{(W-II)},
\end{equation}
using two constants $0\leq c_1<c_2\leq 1$ chosen by the designer of the line search, not the user. W-I is the \emph{Armijo~\cite{armijo1966minimization}}, or \emph{sufficient decrease} condition. It encodes that acceptable functions values should lie below a linear extrapolation line of slope $c_1f'(0)$.  W-II is the \emph{curvature condition}, demanding a decrease in slope. The choice $c_1=0$ accepts any value below $f(0)$, while $c_1=1$ rejects all points for convex functions. For the curvature condition, $c_2=0$ only accepts points with $f'(t)\geq 0$; while $c_2=1$ accepts any point of greater slope than $f'(0)$. W-I and W-II are known as the \emph{weak} form of the Wolfe conditions. The \emph{strong} form replaces W-II with $|f'(t)|\leq c_2 |f'(0)|$ (W-IIa).  This guards against accepting points of low function value but large positive gradient. Figure \ref{fig:minimize-sketch} shows a conceptual sketch illustrating the typical process of a line search, and the weak and strong Wolfe conditions. The exposition in \textsection\ref{sec:determ-conv} will initially focus on the weak conditions, which can be precisely modeled probabilistically. Section \ref{sec:appr-strong-wolfe} then adds an approximate treatment of the strong form.

\subsection{Bayesian Optimization}
\label{sec:bayes-optim}

A recently blossoming sample-efficient approach to global optimization revolves around modeling the objective $f$ with a probability measure $p(f)$; usually a Gaussian process (\gp). Searching for extrema, evaluation points are then chosen by a utility functional $u[p(f)]$. Our line search borrows the idea of a Gaussian process surrogate, and a popular utility, \emph{expected improvement} \citep{jones1998efficient}. Bayesian optimization methods are often computationally expensive, thus ill-suited for a cost-sensitive task like a line search. But since line searches are governors more than information extractors, the kind of sample-efficiency expected of a Bayesian optimizer is not needed. The following sections develop a lightweight algorithm which adds only minor computational overhead to stochastic optimization.

\section{A Probabilistic Line Search}
\label{sec:method}

We now consider minimizing $y(t)=\hat{\mathcal{L}} (x(t))$ from Eq.~(\ref{eq:1}). That is, the algorithm can access only noisy function values and gradients $y_t,y' _t$ at location $t$, with Gaussian likelihood
\begin{equation}
  \label{eq:3}
  p(y _t,y' _t\g f) = \N\left(
    \begin{bmatrix}
      y _t\\ y' _t
    \end{bmatrix};
    \begin{bmatrix}
      f(t) \\ f'(t)
    \end{bmatrix},
    \begin{bmatrix}
      \sigma_{f}^2 & 0 \\ 0 & \sigma_{f'} ^2
    \end{bmatrix}
\right).
\end{equation}
The Gaussian form is supported by the Central Limit argument at Eq.~(\ref{eq:1}), see \textsection\ref{sec:hyperp-estim} regarding estimation of the variances $\sigma^2 _f,\sigma^2 _{f'}$. Our algorithm has three main ingredients: A robust yet lightweight Gaussian process surrogate on $f(t)$ facilitating analytic optimization; a simple Bayesian optimization objective for exploration; and a probabilistic formulation of the Wolfe conditions as a termination criterion.

\subsection{Lightweight Gaussian Process Surrogate}
\label{sec:gauss-proc-surr}

We model information about the objective in a probability measure $p(f)$. There are two requirements on such a measure: First, it must be robust to irregularity of the objective. And second, it must allow analytic computation of discrete candidate points for evaluation, because a line search should not call yet another optimization subroutine itself. Both requirements are fulfilled by a once-integrated Wiener process, i.e.~a zero-mean Gaussian process prior $p(f)=\GP(f;0,k)$ with covariance function
\begin{equation}
  \label{eq:2}
  k(t,t') = \theta^2 \left[\nicefrac{1}{3}\operatorname{min}^3(\tilde{t},\tilde{t}') +
    \nicefrac{1}{2}|t-t'|\operatorname{min}^2(\tilde{t},\tilde{t}') \right].
\end{equation}%
Here $\tilde{t}:=t+\tau$ and $\tilde{t}':=t'+\tau$ denote a shift by a constant $\tau>0$. This ensures this kernel is positive semi-definite, the precise value $\tau$ is irrelevant as the algorithm only considers positive values of $t$ (our implementation uses $\tau=10$). See \textsection\ref{sec:hyperp-estim} regarding the scale $\theta^2$. With the likelihood of Eq.~(\ref{eq:3}), this prior gives rise to a \gp~posterior whose mean function is a cubic spline\footnote{Eq.~(\ref{eq:2}) can be generalized to the `natural spline', removing the need for the constant $\tau$ \citep[\textsection 6.3.1]{RasmussenWilliams}. However, this notion is ill-defined in the case of a single observation, which is crucial for the line search.} \cite{wahba1990spline}. We note in passing that regression on $f$ and $f'$ from $N$ observations of pairs $(y_t,y'_t)$ can be formulated as a filter \cite{sarkka2013bayesian} and thus performed in $\mathcal{O}(N)$ time. However, since a line search typically collects $<10$ data points, generic \gp~inference, using a Gram matrix, has virtually the same, low cost.

Because Gaussian measures are closed under linear maps \citep[\textsection 10]{papoulis91:probab_random}, Eq.~(\ref{eq:2}) implies a Wiener process (linear spline) model on $f'$:
\begin{equation}
  \label{eq:6}
    p(f;f') = \GP\left(
      \begin{bmatrix}
        f \\ f'
      \end{bmatrix}; \vec{0},
      \begin{bmatrix}
        k & \kd\\ \dk & \dkd
      \end{bmatrix}
    \right),
\end{equation}
with (using the indicator function $\II(x)=1$ if $x$, else 0)
\begin{equation}
 \label{eq:23}
\tensor[^{\de^i}]{k}{^{\de^j}} = \frac{\de^{i+j}k(t,t')}{\de t^i \de {t'}^j},
\q\text{thus}\q
\begin{array}{rl}
    \kd(t,t') =& \theta^2 \left[ \II(t<t') \nicefrac{t^2}{2} + \II(t\geq t') (tt' - \nicefrac{{t'}^2}{2}) \right]\\
    \dk(t,t') =& \theta^2 \left[ \II(t'<t) \nicefrac{t'^2}{2} + \II(t'\geq t) (tt' - \nicefrac{{t}^2}{2}) \right]\\
    \dkd(t,t')=& \theta^2 \min(t,t')
\end{array}.
\end{equation}
Given a set of evaluations $(\vec{t},\y,\y')$ (vectors, with elements $t_i,y_{t_i},y'_{t_i}$) with independent likelihood (\ref{eq:3}), the posterior $p(f\g \y,\y')$ is a \gp~with posterior mean $\mu$ and covariance and $\tilde{k}$ as follows:
\begin{align}
  \label{eq:7}
  \mu(t) &=
  \underbrace{\begin{bmatrix}
    k_{\vec{t}t} \\ \dk_{\vec{t}t}
  \end{bmatrix}\Trans
  \left(
    \begin{bmatrix}
      k_{\vec{t}\vec{t}}+\sigma_f ^2 \Id & \kd_{\vec{t}\vec{t}}\\
      \dk_{\vec{t}\vec{t}} & \dkd_{\vec{t}\vec{t}} +\sigma_{f'} ^2 \Id
    \end{bmatrix}\right)^{-1}
}_{=:\vec{g}\Trans(t)}
  \begin{bmatrix}
    \y \\ \y'
  \end{bmatrix},\q
  \tilde{k}(t,t') = k_{tt'} - 
  \vec{g}\Trans(t)
  \begin{bmatrix}
    k_{\vec{t} t'} \\ \dk_{\vec{t}t'}
  \end{bmatrix}.
\end{align}
The posterior marginal variance will be denoted by $\V(t)=\tilde{k}(t,t)$. To see that $\mu$ is indeed piecewise cubic (i.e.~a cubic spline), we note that it has at most three non-vanishing derivatives\footnote{There is no well-defined probabilistic belief over $f''$ and higher derivatives---sample paths of the Wiener process are almost surely non-differentiable almost everywhere \citep[\textsection 2.2]{adler1981geometry}. But $\mu(t)$ is always a member of the reproducing kernel Hilbert space induced by $k$, thus piecewise cubic \citep[\textsection 6.1]{RasmussenWilliams}.}, because
\begin{xalignat}{2}
  \notag%
  \ddk(t,t') &= \theta^2 \II(t\leq t')(t' - t) &
  \ddkd(t,t') &= \theta^2 \II(t\leq t')\\
\label{eq:9}
  \dddk(t,t') &= -\theta^2 \II(t\leq t') &
  \dddkd(t,t') &= 0.
\end{xalignat}
This piecewise cubic form of $\mu$ is crucial for our purposes: having collected $N$ values of $f$ and $f'$, respectively, all local minima of $\mu$ can be found analytically in $\mathcal{O}(N)$ time in a single sweep through the `cells' $t_{i-1}<t<t_{i}$, $i=1,\dots,N$ (here $t_0=0$ denotes the start location, where $(y_0,y'_0)$ are `inherited' from the preceding line search. For typical line searches $N<10$, c.f.~\textsection\ref{sec:experiments}). In each cell, $\mu(t)$ is a cubic polynomial with at most one minimum in the cell, found by a trivial quadratic computation from the three scalars $\mu'(t_i),\mu''(t_i),\mu'''(t_i)$. This is in contrast to other \gp~regression models---for example the one arising from a Gaussian kernel---which give more involved posterior means whose local minima can be found only approximately. Another advantage of the cubic spline interpolant is that it does not assume the existence of higher derivatives (in contrast to the Gaussian kernel, for example), and thus reacts robustly to irregularities in the objective. 

In our algorithm, after each evaluation of $(y_N,y'_N)$, we use this property to compute a short list of \emph{candidates} for the next evaluation, consisting of the $\leq N$ local minimizers of $\mu(t)$ and one additional \emph{extrapolation} node at $t_{\max}+\alpha$, where $t_{\max}$ is the currently largest evaluated $t$, and $\alpha$ is an extrapolation step size starting at $\alpha=1$ and doubled after each extrapolation step. 

\subsection{Choosing Among Candidates}
\label{sec:select-eval-points}

The previous section described the construction of $<N+1$ discrete candidate points for the next evaluation. To decide at which of the candidate points to actually call $f$ and $f'$, we make use of a popular utility from Bayesian optimization. \emph{Expected improvement}~\cite{jones1998efficient} is the expected amount, under the \gp~surrogate, by which the function $f(t)$ might be smaller than a `current best' value $\eta$ (we set $\eta=\min_{i=0,\dots,N}\{\mu(t_i)\}$, where $t_i$ are observed locations),
\begin{equation}
  \label{eq:10}
  \begin{split}
    u&_{\text{EI}}(t) = \Exp_{p(f_t\g \vec{y},\vec{y}')}[\min\{0,\eta -
    f(t)\}]\\
    &= \frac{\eta - \mu(t)}{2}\left(1+\erf\frac{\eta-\mu(t)}{\sqrt{2\V(t)}}\right) +
    \sqrt{\frac{\V(t)}{2\pi}}\exp\left(-\frac{(\eta-\mu(t))^2}{2\V(t)} \right).
  \end{split}
\end{equation}
The next evaluation point is chosen as the candidate maximizing this utility, multiplied by the probability for the Wolfe conditions to be fulfilled, which is derived in the following section.

\begin{figure*}
  \centering
  \setlength{\figwidth}{.13\textwidth}
  \setlength{\figheight}{.1\textheight}
  {\scriptsize %
    \includegraphics{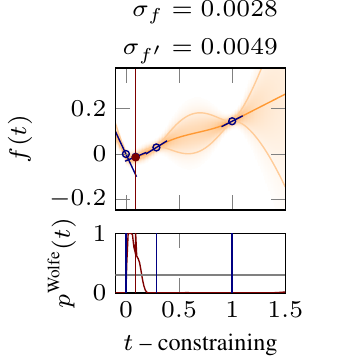}\hspace{-7mm}%
    \includegraphics{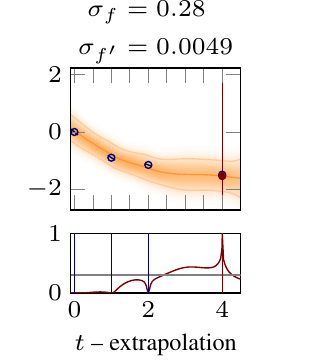}\hspace{-7mm}%
    \includegraphics{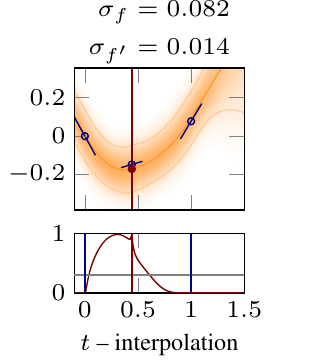}\hspace{-7mm}%
    \includegraphics{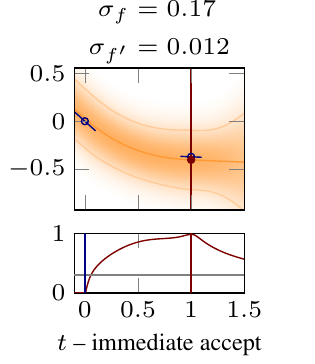}\hspace{-7mm}%
    \includegraphics{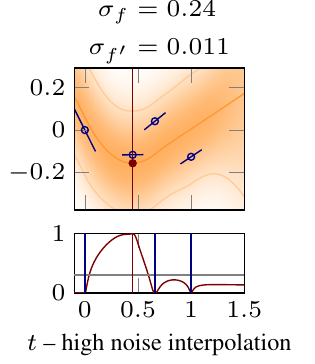}%
  }
  \caption{Curated snapshots of line searches (from MNIST experiment, \textsection\ref{sec:experiments}), showing variability of the objective's shape and the decision process. Top row: \gp~posterior and evaluations, bottom row: approximate $\pW$ over strong Wolfe conditions. Accepted point marked red.}
\label{fig:snapshots}
\end{figure*}

\subsection{Probabilistic Wolfe Conditions for Termination}
\label{sec:determ-conv}

The key observation for a probabilistic extension of W-I and W-II is that they are positivity constraints on two variables $a_t,b_t$ that are both linear projections of the (jointly Gaussian) variables $f$ and $f'$:
\begin{equation}
  \label{eq:11}
  \begin{bmatrix}
    a_t\\b_t
  \end{bmatrix} = 
  \begin{bmatrix}
    1 & c_1t & -1 & 0\\
    0 & -c_2 & 0 & 1
  \end{bmatrix}
  \begin{bmatrix}
    f(0)\\ f'(0) \\ f(t) \\ f'(t)
  \end{bmatrix} \geq 0.
\end{equation}
The \gp~of Eq.~\eqref{eq:6}~on $f$ thus implies, at each value of $t$, a bivariate Gaussian distribution
\begin{align}
  \label{eq:12}
  p(a_t,b_t) &= \N\left(
    \begin{bmatrix}
      a_t\\b_t
    \end{bmatrix};
    \begin{bmatrix}
      m^a _t\\m^b _t
    \end{bmatrix},
    \begin{bmatrix}
      C^{aa} _{t} & C^{ab} _t\\ C^{ba} _t & C^{bb} _t
    \end{bmatrix}\right),\\
    \label{eq:13}
    \text{with}\qq m^a _t &= \mu(0) - \mu(t) + c_1 t \mu'(0)\qq\text{and}\qq
    m^b _t = \mu'(t) - c_2 \mu'(0)\\
  \begin{split}
  \label{eq:14}
    \text{and}\qq C^{aa} _t &= \tilde{k}_{00} + (c_1t)^2 \tensor*[^\de]{\tilde{k}}{^\de _{00}}
    + \tilde{k}_{tt}
    + 2[c_1t(\tilde{k}^{\de} _{00} - \tensor[^\de]{\tilde{k}}{_{0t}}) - \tilde{k}_{0t}]\\
    C^{bb} _t &= c_2 ^2 \tensor*[^\de]{\tilde{k}}{^\de _{00}} - 2c_2
    \tensor*[^{\de}]{\tilde{k}}{^{\de}_{0t}} + \tensor*[^{\de}]{\tilde{k}}{^{\de} _{tt}}\\
    C^{ab} _t = C^{ba} _t &= -c_2(\tilde{k}^{\de} _{00} + c_1 t
    \tensor*[^{\de}]{\tilde{k}}{^{\de} _{00}})
    + c_2 \tensor[^{\de}]{\tilde{k}}{_{0t}}+ \tensor[^{\de}]{\tilde{k}}{_{t0}} + c_1t
    \tensor[^{\de}]{\tilde{k}}{^{\de} _{0t}} - \tilde{k}^{\de} _{tt}.
  \end{split}
\end{align}
The quadrant probability $\pW_t=p(a_t>0 \wedge b_t>0)$ for the Wolfe conditions to hold is an integral over a bivariate normal probability,
\begin{equation}
  \label{eq:16}
  \pW_t = 
  \int_{-\frac{m^a _t}{\sqrt{C^{aa} _t}}} ^\infty
  \int_{-\frac{m^b _t}{\sqrt{C^{bb} _t}}}
  ^\infty \N\left(
    \begin{bmatrix}
      a\\b
    \end{bmatrix};
    \begin{bmatrix}
      0\\0
    \end{bmatrix}
    ,
    \begin{bmatrix}
      1 & \rho_t \\ \rho_t & 1
    \end{bmatrix}
\right) \,da \,db,
\end{equation}
with correlation coefficient $\rho_t = C^{ab} _t / \sqrt{C^{aa} _t C^{bb} _t}$. It can be computed efficiently~\cite{drezner1990computation}, using readily available code\footnote{e.g.~{\scriptsize\url{http://www.math.wsu.edu/faculty/genz/software/matlab/bvn.m}}} (on a laptop, one evaluation of $\pW_t$ cost about 100 microseconds, each line search requires $<50$ such calls). The line search computes this probability for all evaluation nodes, after each evaluation. If any of the nodes fulfills the Wolfe conditions with $\pW_t>c_W$, greater than some threshold $0<c_W\leq 1$, it is accepted and returned. If several nodes simultaneously fulfill this requirement, the $t$ of the lowest $\mu(t)$ is returned. Section~\ref{sec:design-param-c_1} below motivates fixing $c_W=0.3$.

\subsubsection{Approximation for strong conditions:}
\label{sec:appr-strong-wolfe}

As noted in Section~\ref{sec:wolfe-cond-conv}, deterministic optimizers tend to use the strong Wolfe conditions, which use $|f'(0)|$ and $|f'(t)|$. A precise extension of these conditions to the probabilistic setting is numerically taxing, because the distribution over $|f'|$ is a non-central $\chi$-distribution, requiring customized computations. However, a straightforward variation to (\ref{eq:16}) captures the spirit of the strong Wolfe conditions, that large positive derivatives should not be accepted: Assuming $f'(0)<0$ (i.e.~that the search direction is a descent direction), the strong second Wolfe condition can be written exactly as
\begin{equation}
  \label{eq:18}
  0 \leq b_t = f'(t)-c_2f(0) \leq -2c_2 f'(0).
\end{equation}
The value $-2c_2 f'(0)$ is bounded to $95\%$ confidence by
\begin{equation}
  \label{eq:19}
  -2c_2 f'(0)  \lesssim -2 c_2 (|\mu'(0)| + 2\sqrt{\V'(0)}) =:\bar{b}.
\end{equation}
Hence, an approximation to the strong Wolfe conditions can be reached by replacing the infinite upper integration limit on $b$ in Eq.~(\ref{eq:16}) with $(\bar{b}-m^b _t)/\sqrt{C^{bb} _t}$. The effect of this adaptation, which adds no overhead to the computation, is shown in Figure~\ref{fig:prob_ls_sketch} as a dashed line.

\subsection{Eliminating Hyper-parameters}
\label{sec:hyperp-estim}

As a black-box inner loop, the line search should not require any tuning by the user. The preceding section introduced six so-far undefined parameters: $c_1,c_2,c_W,\theta,\sigma_f,\sigma_{f'}$. We will now show that $c_1,c_2,c_W$, can be fixed by hard design decisions. $\theta$ can be eliminated by standardizing the optimization objective within the line search; and the noise levels can be estimated at runtime with low overhead for batch objectives of the form in Eq.~(\ref{eq:1}). The result is a parameter-free algorithm that effectively \emph{removes} the one most problematic parameter from \sgd---the learning rate.

\paragraph{Design Parameters $c_1,c_2,c_W$}
\label{sec:design-param-c_1}

Our algorithm inherits the Wolfe thresholds $c_1$ and $c_2$ from its deterministic ancestors. We set $c_1=0.05$ and $c_2 = 0.8$. This is a standard setting that yields a `lenient' line search, i.e.~one that accepts most descent points. The rationale is that the stochastic aspect of \sgd~is not always problematic, but can also be helpful through a kind of `annealing' effect.

The acceptance threshold $c_W$ is a new design parameter arising only in the probabilistic setting. We fix it to $c_W=0.3$. To motivate this value, first note that in the noise-free limit, all values $0<c_W<1$ are equivalent, because $\pW$ then switches discretely between 0 and 1 upon observation of the function. A back-of-the-envelope computation (left out for space), assuming only two evaluations at $t=0$ and $t=t_1$ and the same fixed noise level on $f$ and $f'$ (which then cancels out), shows that function values barely fulfilling the conditions, i.e.~$a_{t_1}=b_{t_1}=0$, can have $\pW\sim 0.2$ while function values at $a_{t_1}=b_{t_1}=-\eps$ for $\eps\to 0$ with `unlucky' evaluations (both function and gradient values one standard-deviation from true value) can achieve $\pW\sim 0.4$. The choice $c_W=0.3$ balances the two competing desiderata for precision and recall. Empirically (Fig.~\ref{fig:snapshots}), we rarely observed values of $\pW$ close to this threshold. Even at high evaluation noise, a function evaluation typically either clearly rules out the Wolfe conditions, or lifts $\pW$ well above the threshold. 

\paragraph{Scale $\theta$}
\label{sec:scale-theta}

The parameter $\theta$ of Eq.~(\ref{eq:2}) simply scales the prior variance. It can be eliminated by scaling the optimization objective: We set $\theta=1$ and scale
  $y_i \gets \nicefrac{(y_i - y_0)}{|y' _0|}, y_i' \gets \nicefrac{y'_i}{|y' _0|}$ within the code of the line search. This gives $y(0)=0$ and $y'(0)=-1$, and typically ensures the objective ranges in the single digits across $0<t<10$, where most line searches take place. The division by $|y' _0|$ causes a non-Gaussian disturbance, but this does not seem to have notable empirical effect.

\paragraph{Noise Scales $\sigma_f,\sigma_{f'}$}
\label{sec:noise-variances}

The likelihood (\ref{eq:3}) requires standard deviations for the noise on both function values ($\sigma_f$) and gradients ($\sigma_{f'}$).  One could attempt to learn these across several line searches. However, in exchangeable models, as captured by Eq.~(\ref{eq:1}), the variance of the loss and its gradient can be estimated directly within the batch, at low computational overhead---an approach already advocated by Schaul et al.~\cite{schaul2013no}. We collect the empirical statistics
\begin{equation}
  \label{eq:21}
    \hat{S}(x) := \frac{1}{m}\sum_{j} ^m \ell^2(x,y_j),\qq\text{and} \qq
    \hat{\nabla S}(x):= \frac{1}{m}\sum_{j} ^m \nabla\ell(x,y_j).^2 
\end{equation}
(where $.^2$ denotes the element-wise square) and estimate, at the beginning of a line search from $x_k$,
\begin{equation}
  \label{eq:22}
    \sigma_f ^2=\frac{1}{m-1}\left(\hat{S}(x_k) - \hat{\mathcal{L}}(x_k)^2\right)\qq\text{and}\qq
    \sigma_{f'} ^2={s_i.^2}\Trans\left[\frac{1}{m-1}\left(\hat{\nabla S}(x_k) - (\nabla\hat{\mathcal{L}}).^2\right)\right].
\end{equation}
This amounts to the cautious assumption that noise on the gradient is independent. We finally scale the two empirical estimates as described in \textsection\ref{sec:scale-theta}: $\sigma_f \gets \sigma_f / |y'(0)|$, and ditto for $\sigma_{f'}$. The overhead of this estimation is small if the computation of $\ell(x,y_j)$ itself is more expensive than the summation over $j$ (in the neural network examples of \textsection\ref{sec:experiments}, with their comparably simple $\ell$, the additional steps added only $\sim 1\%$ cost overhead to the evaluation of the loss). Of course, this approach requires a batch size $m>1$. For single-sample batches, a running averaging could be used instead (single-sample batches are not necessarily a good choice. In our experiments, for example, vanilla \sgd~with batch size 10 converged faster in wall-clock time than unit-batch \sgd). Estimating noise separately for each input dimension captures the often inhomogeneous structure among gradient elements, and its effect on the noise along the projected direction. For example, in deep models, gradient noise is typically higher on weights between the input and first hidden layer, hence line searches along the corresponding directions are noisier than those along directions affecting higher-level weights.

\subsubsection{Propagating Step Sizes Between Line Searches}
\label{sec:step-size-alpha}

As will be demonstrated in \textsection\ref{sec:experiments}, the line search can find good step sizes even if the length of the direction $s_i$ (which is proportional to the learning rate $\alpha$ in \sgd) is mis-scaled. Since such scale issues typically persist over time, it would be wasteful to have the algorithm re-fit a good scale in each line search. Instead, we propagate step lengths from one iteration of the search to another: We set the initial search direction to $s_0=-\alpha_0 \nabla \hat{\mathcal{L}} (x_0)$ with some initial learning rate $\alpha_0$. Then, after each line search ending at $x_i=x_{i-1} + t_* s_i$, the next search direction is set to $s_{i+1}=-1.3\cdot t_* \alpha_0 \nabla \hat{\mathcal{L}} (x_i)$. Thus, the next line search starts its extrapolation at $1.3$ times the step size of its predecessor.

\paragraph{Remark on convergence of \sgd~with line searches:} We note in passing that it is straightforward to ensure that \sgd~instances using the line search inherit the convergence guarantees of \sgd: Putting even an extremely loose bound $\bar{\alpha}_i$ on the step sizes taken by the $i$-th line search, such that $\sum_i ^\infty \bar{\alpha}_i = \infty$ and $\sum_i ^\infty \bar{\alpha}_i ^2 <\infty$, ensures the line search-controlled \sgd~converges in probability~\cite{robbins1951stochastic}.

\section{Experiments}
\label{sec:experiments}

Our experiments were performed on the well-worn problems of training a 2-layer neural net with logistic nonlinearity on the MNIST and CIFAR-10 datasets.\footnote{\scriptsize \url{http://yann.lecun.com/exdb/mnist/} and \url{http://www.cs.toronto.edu/~kriz/cifar.html}. Like other authors, we only used the ``batch 1'' sub-set of CIFAR-10.} In both cases, the network had 800 hidden units, giving optimization problems with $636\,010$ and $2\,466\,410$ parameters, respectively. While this may be `low-dimensional' by contemporary standards, it exhibits the stereotypical challenges of stochastic optimization for machine learning. Since the line search deals with only univariate subproblems, the extrinsic dimensionality of the optimization task is not particularly relevant for an empirical evaluation. Leaving aside the cost of the function evaluations themselves, computation cost associated with the line search is independent of the extrinsic dimensionality.

\begin{figure}
  \centering
  \setlength{\figwidth}{.41\textwidth}
  \setlength{\figheight}{.1\textheight}
  {
  {\scriptsize%
   \includegraphics{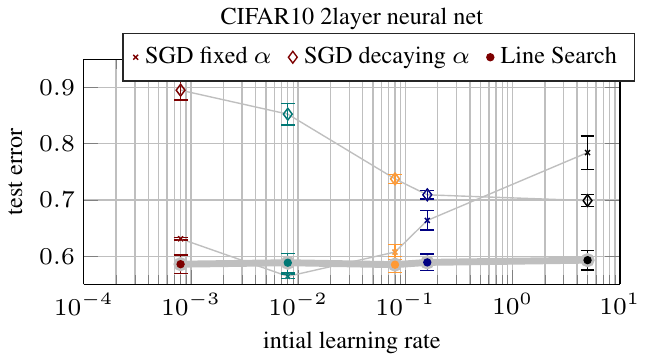}
   \includegraphics{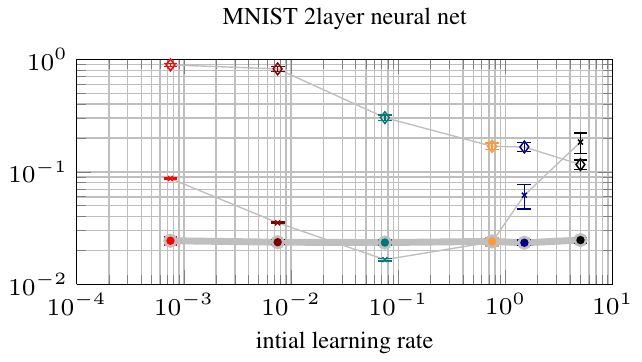}\\
   \includegraphics{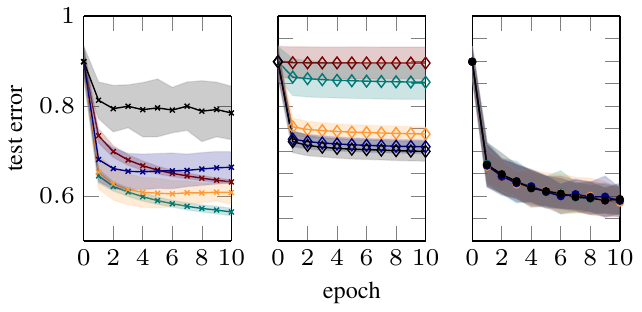}
   \includegraphics{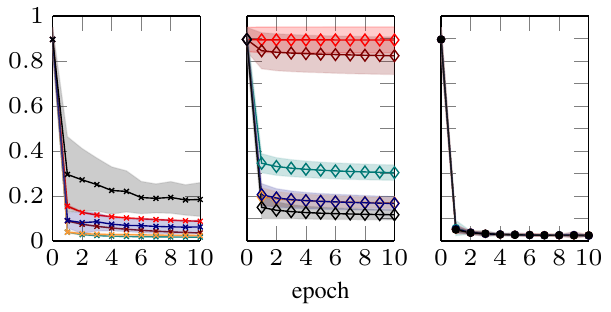}\\}
  }
  \caption{Top row: test error after $10$ epochs as function of initial learning rate (note logarithmic ordinate for MNIST). Bottom row: Test error as function of training epoch (same color and symbol scheme as in top row). No matter the initial learning rate, the line search-controlled \sgd~perform close to the (in practice unknown) optimal \sgd~instance, effectively removing the need for exploratory experiments and learning-rate tuning. All plots show means and 2 std.-deviations over 20 repetitions.}
\label{fig:alpha-sweep}
\end{figure}

The central nuisance of \sgd~is having to choose the learning rate $\alpha$, and potentially also a schedule for its decrease. Theoretically, a decaying learning rate is necessary to guarantee convergence of \sgd~\cite{robbins1951stochastic}, but empirically, keeping the rate constant, or only decaying it cautiously, often work better (Fig.~\ref{fig:alpha-sweep}). In a practical setting, a user would perform exploratory experiments (say, for $10^3$ steps), to determine a good learning rate and decay schedule, then run a longer experiment in the best found setting. In our networks, constant learning rates of $\alpha=0.75$ and $\alpha=0.08$ for MNIST and CIFAR-10, respectively, achieved the lowest test error after the first $10^3$ steps of \sgd. 
We then trained networks with vanilla \sgd~with and without $\alpha$-decay (using the schedule $\alpha(i)=\alpha_0/i$), and \sgd~using the probabilistic line search, with $\alpha_0$ ranging across five orders of magnitude, on batches of size $m=10$. 

Fig.~\ref{fig:alpha-sweep}, top, shows test errors after 10 epochs as a function of the initial learning rate $\alpha_0$ (error bars based on 20 random re-starts). Across the broad range of $\alpha_0$ values, the line search quickly identified good step sizes $\alpha(t)$, stabilized the training, and progressed efficiently, reaching test errors similar to those reported in the literature for tuned versions of this kind of architecture on these datasets. While in both datasets, the best \sgd~instance without rate-decay just barely outperformed the line searches, the optimal $\alpha$ value was \emph{not} the one that performed best after $10^3$ steps. So this kind of exploratory experiment (which comes with its own cost of human designer time) would have led to worse performance than simply starting a single instance of \sgd~with the linesearch and $\alpha_0=1$, letting the algorithm do the rest.

Average time overhead (i.e.~excluding evaluation-time for the objective) was about 48ms per line search. This is \emph{independent} of the problem dimensionality, and expected to drop significantly with optimized code. Analysing one of the MNIST instances more closely, we found that the average length of a line search was $\sim 1.4$ function evaluations, $80\%-90\%$ of line searches terminated after the first evaluation. This suggests good scale adaptation and thus efficient search (note that an `optimally tuned' algorithm would always lead to accepts). 



The supplements provide additional plots, of raw objective values, chosen step-sizes, encountered gradient norms and gradient noises during the optimization, as well as test-vs-train error plots, for each of the two datasets, respectively. These provide a richer picture of the step-size control performed by the line search. In particular, they show that the line search chooses step sizes that follow a nontrivial dynamic over time. This is in line with the empirical truism that \sgd~requires tuning of the step size during its progress, a nuisance taken care of by the line search. Using this structured information for more elaborate analytical purposes, in particular for convergence estimation, is an enticing prospect, but beyond the scope of this paper.

\section{Conclusion}
\label{sec:conclusion}

The line search paradigm widely accepted in deterministic optimization can be extended to noisy settings. Our design combines existing principles from the noise-free case with ideas from Bayesian optimization, adapted for efficiency. We arrived at a lightweight ``black-box'' algorithm that exposes no parameters to the user. Our method is complementary to, and can in principle be combined with, virtually all existing methods for stochastic optimization that adapt a step \emph{direction} of fixed \emph{length}. Empirical evaluations suggest the line search effectively frees users from worries about the choice of a learning rate: Any reasonable initial choice will be quickly adapted and lead to close to optimal performance. Our matlab implementation can be found at \scriptsize\url{http://tinyurl.com/probLineSearch}. 

\clearpage
\small 
\bibliographystyle{unsrtnat}
\bibliography{./../../../../bibfile}
\clearpage
\normalsize


\setcounter{section}{0}
\setcounter{figure}{0}
\begin{center}
\huge{\textbf{Supplements}}
\end{center}

This supplementary document contains additional results of the experiments described in the main paper, using the probabilistic line search algorithm to control the learning rate in stochastic gradient descent during training of two-layer neural network architectures in the CIFAR-10 and MNIST datasets. 

\section{Evolution of Function Values}

\begin{figure}[b]
  \centering
  \setlength{\figwidth}{.4\textwidth}
  \setlength{\figheight}{.25\textheight}
{
  {\scriptsize%
   \includegraphics{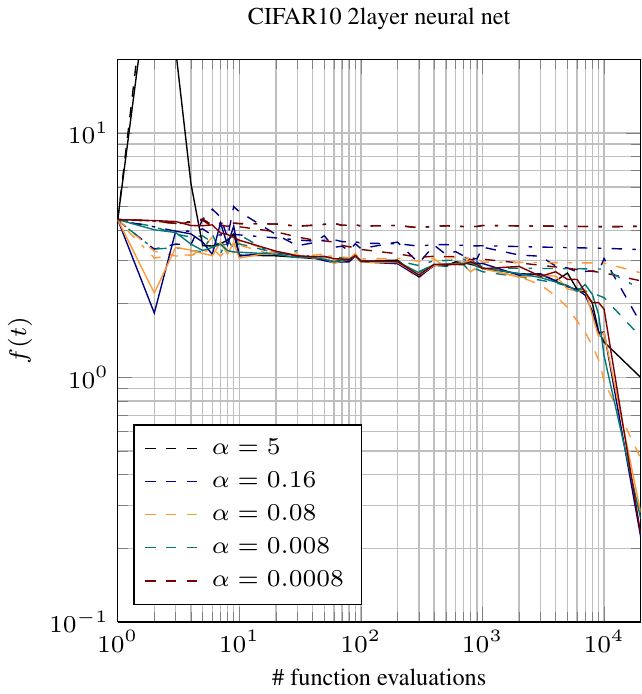}
   \includegraphics{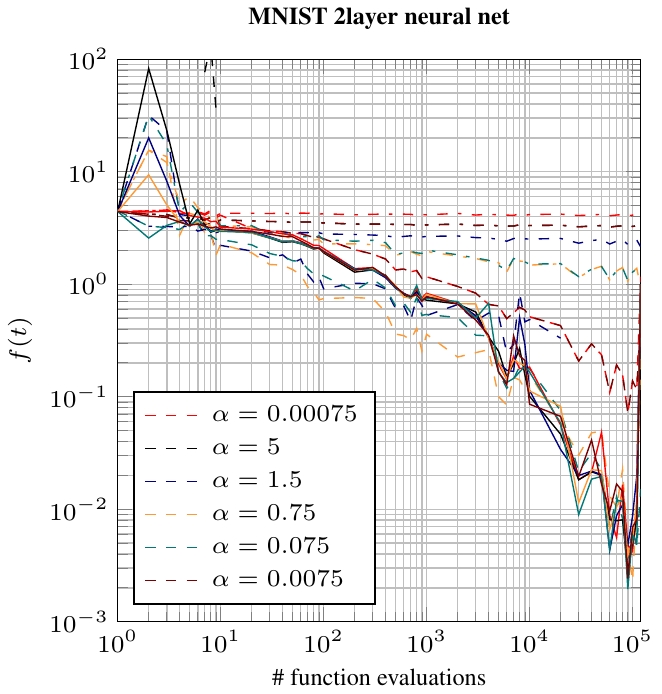}}
   }
  \caption{Function values encountered during neural network experiments. Vanilla \sgd~at different fixed learning rate $\alpha$ as dashed lines, \sgd~at different decaying learning rates as dashed-dotted lines. Solid lines show results using the probabilistic line search initialized at the corresponding $\alpha$-values.}
  \label{fig:func-values}
\end{figure}

Figure \ref{fig:func-values} plots the evolution of encountered raw function values against function evaluations. Each function call evaluates the gradient on a batch of size $10$, both for \sgd~with constant and decaying learning rate, and for the line search-enhanced \sgd. To keep the plot readable, the plot lines have been smoothed with a windowed running average, and only plotted at logarithmically spaced points. Among the noteworthy features of these plots is that \sgd~with large step sizes can be unstable (divergent dashed black lines), while this instability is caught and controlled by the line search. Regardless of initial step size, all line search-controlled instances perform very similarly, and reach close to optimal performance. Over the dynamic development of the optimization process, some specific choices of step size temporarily perform better than the line search-controlled instances, but this advantage slims or vanishes over time, because no fixed step size is optimal over the course of the entire optimization process. As mentioned in the main paper, finding those optimal step sizes would normally involve a tedious, costly search, which the line search effecively removes.

\section{Optimal Step Sizes Vary During Training}
\label{sec:optimal-step-sizes}

\begin{figure}
  \centering
  \setlength{\figwidth}{.4\textwidth}
  \setlength{\figheight}{.33\textheight}
  {
  {\scriptsize%
   \includegraphics{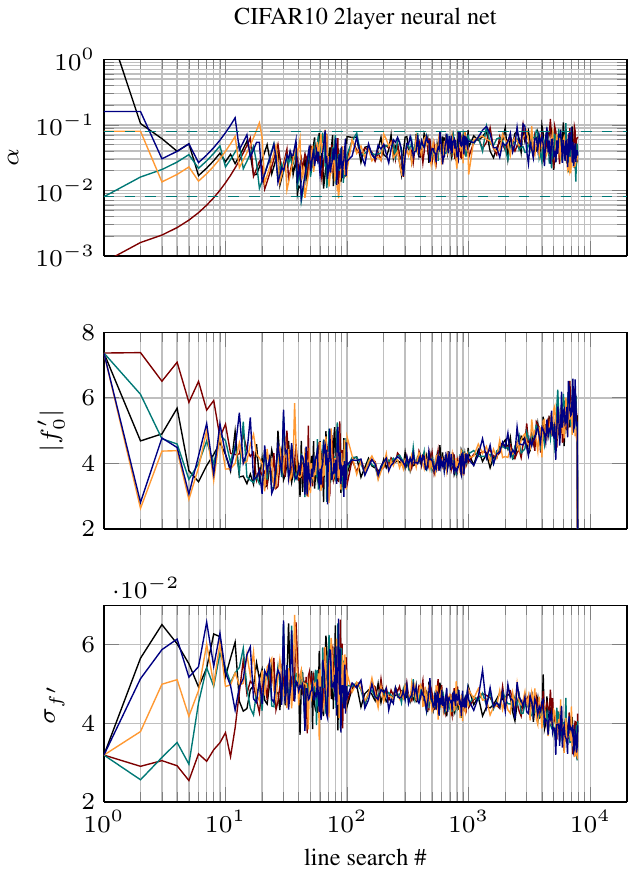}
   \includegraphics{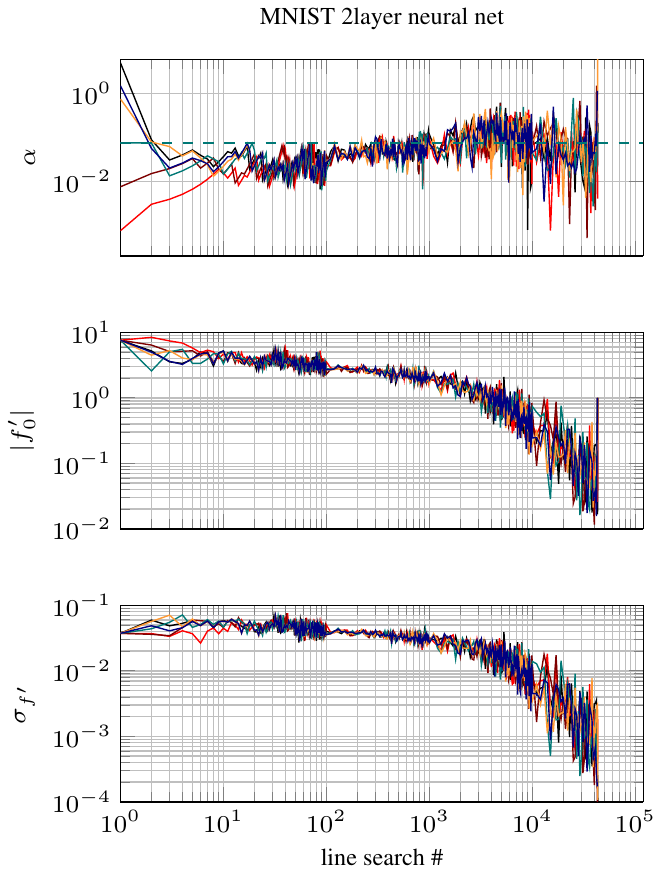}}
   }
  \caption{Same colors as Fig.~4 in the main paper: Accepted step sizes $\alpha$, initial gradients $|f'_0|$ and absolute noise on gradient $\sigma_{f'}$. The probabilistic line search quickly fixes even wildly varying initial learning rates, within the first few line searches. The accepted step lengths drift over the course of the optimization, in a nontrivial relation to noise levels and gradient norms.}
  \label{fig:alpha-trace}
\end{figure}

It is a ``widely known empirical fact'' that \sgd~instances require a certain amount of run-time ``tweaking'', because the optimal step size depends not just on the local structure of the objective, but also on batch-size (and associatd noise level). Figure \ref{fig:alpha-trace} shows accepted step sizes, initial gradients at each search, and estimated gradient noise levels for the line search instances in the same experimental runs described above (smoothed and thinned as in Fig.~\ref{fig:func-values} above). Starting five orders of magnitude apart, the line searches very quickly converged to similar step sizes; and indeed eventually settle around the empirically optimal value of $\alpha=0.075$ (MNIST) and $\alpha = 0.08$ (CIFAR-10) (dashed green horizontal line in Fig.~\ref{fig:alpha-trace}). But step sizes varied over time: starting out small, they then \emph{increased}, and began decreasing again after around $10^4$ line searches. This corroborates the empirical truism that learning rates should not immediately start decreasing, and only do so slowly. Interestingly, while there is an association between gradient values, noise and accepted step sizes, there appears to be no simple analytic relationship between the three. Overall, the emerging picture is that there is indeed nontrivial structure in the objective that is picked up by the line search.

\section{Line Searches Do Not Affect Overfitting}
\begin{figure}
\centering
\setlength{\figwidth}{.4\textwidth}
  \setlength{\figheight}{.2\textheight}
  {
  {\scriptsize%
   \includegraphics{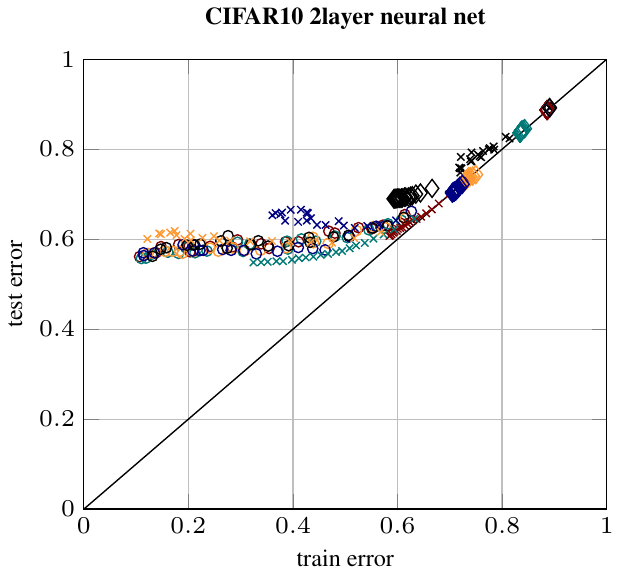}
   \includegraphics{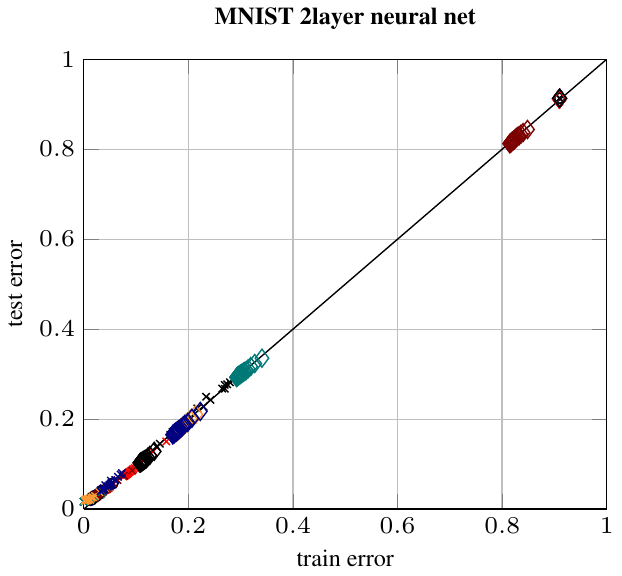}}
   }
   \caption{Test set error rates plotted against training set error rates. Same symbols and colors as in Figure 4 of the main paper. While there is significant over-fitting in CIFAR-10, and virtually no over-fitting in the MNIST case, the line search-controlled instances of \sgd~perform similarly to the best \sgd~instances.}
   \label{fig:test-v-train-error} 
\end{figure}

\begin{figure}
  \centering
  \setlength{\figwidth}{.41\textwidth}
  \setlength{\figheight}{.1\textheight}
  {
  {\scriptsize%
   \includegraphics{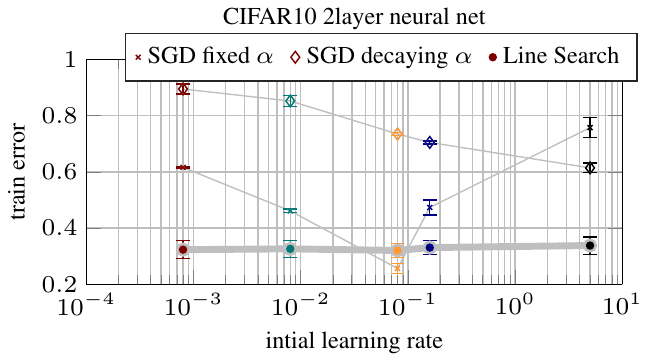}
   \includegraphics{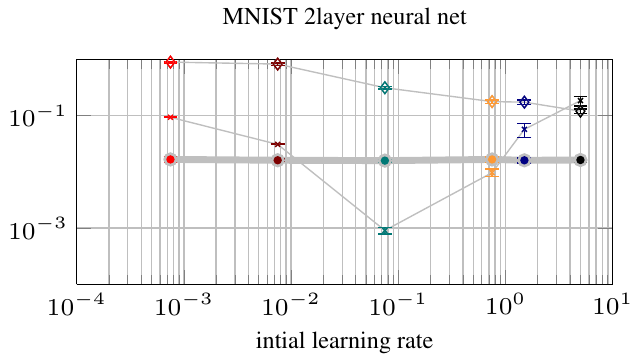}\\
   \includegraphics{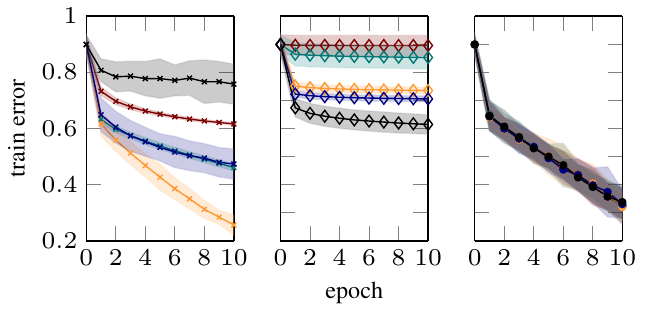}
   \includegraphics{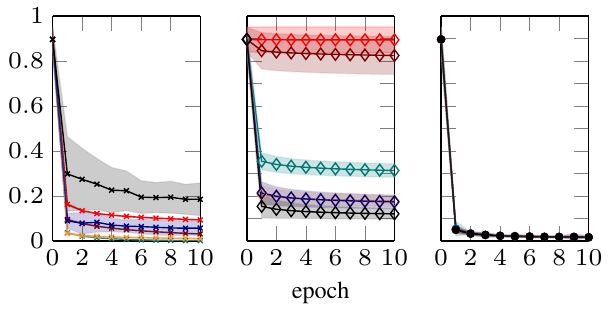}\\}
  }
  \caption{Top row: train error after $10$ epochs as function of initial learning rate (note logarithmic ordinate for MNIST). Bottom row: Train error as function of training epoch (same color and symbol scheme as in top row). Same symbols and colors as in Figure 4 of the main paper. No matter the initial learning rate, the line search-controlled \sgd~perform close to the (in practice unknown) optimal \sgd~instance, effectively removing the need for exploratory experiments and learning-rate tuning. All plots show means and 2 std.-deviations over 20 repetitions.}
\label{fig:alpha-sweep_sup}
\end{figure}

A final worry one might have is that the control interventions of the line search might curtail an ``accidental'' benefical property of \sgd---for example that the somewhat erratic, stochastic steps caused by stochasticity in the gradients allow \sgd~to ``jump over'' local minima of the objective. Such local minima can be a cause of over-fitting, or generally of low empirical performance. The plots in the main paper already confirm that the line search does not cause a stagnation in optimization performance, and can indeed improve this performance drastically. For completeness, Figure \ref{fig:test-v-train-error}  also shows the relation between encountered train- and test-set error rates over the course of the optimization and Figure \ref{fig:alpha-sweep_sup} shows the evolution of the train-set error per epoch as well as its dependence on the initial learning rate (same symbols and colors as in Figure 4 of main paper). There is generally  little over-fitting in MNIST, and fairly strong over-fitting in CIFAR-10. But the instances controlled by the line search (circles) do not show a noticeably different behaviour, in this regard, to the uncontrolled, diffusive \sgd~instances. This suggests that, when the line search intervenes to curtail steps of \sgd, it does so typically to prevent a truly sub-optimal step, rather than a beneficial ``hop'' over the walls surrounding a local minimum.



\end{document}